\documentclass[siggraph]{acmart}

\def\BibTeX{{\rm B\kern-.05em{\sc i\kern-.025em b}\kern-.08emT\kern-.1667em\lower.7ex\hbox{E}\kern-.125emX}}
    
\acmDOI{10.1145/nnnnnnn.nnnnnnn}

\acmISBN{978-x-xxxx-xxxx-x/YY/MM}

\acmConference[Fast-DENSER++]{}{}{}
\acmYear{2019}
\copyrightyear{2019}

\usepackage{algorithm2e}
\usepackage{multirow}
\usepackage{graphicx}
\usepackage{subcaption}

\AtBeginEnvironment{align}{\setcounter{equation}{0}}

\begin{document}
\title[Fast-DENSER++: Evolving Fully-Trained Deep Artificial Neural Networks]{Fast-DENSER++: Evolving Fully-Trained\\Deep Artificial Neural Networks}

\author{Filipe Assun\c{c}\~{a}o, Nuno Louren\c{c}o, Penousal Machado, and Bernardete Ribeiro}
\affiliation{ 
      \institution{CISUC, Department of Informatics Engineering,\\University of Coimbra, Coimbra, Portugal}
    }
\email{{fga,naml,machado,bribeiro}@dei.uc.pt}

\renewcommand{\shortauthors}{F. Assun\c{c}\~{a}o et al.}

\begin{abstract}
This paper proposes a new extension to Deep Evolutionary Network Structured Evolution (DENSER), called Fast-DENSER++ (F-DENSER++). The vast majority of NeuroEvolution methods that optimise Deep Artificial Neural Networks (DANNs) only evaluate the candidate solutions for a fixed amount of epochs; this makes it difficult to effectively assess the learning strategy, and requires the best generated network to be further trained after evolution. F-DENSER++ enables the training time of the candidate solutions to grow continuously as necessary, i.e., in the initial generations the candidate solutions are trained for shorter times, and as generations proceed it is expected that longer training cycles enable better performances. Consequently, the models discovered by F-DENSER++ are fully-trained DANNs, and are ready for deployment after evolution, without the need for further training. The results demonstrate the ability of F-DENSER++ to effectively generate fully-trained DANNs; by the end of evolution, whilst the average performance of the models generated by F-DENSER++ is of 88.73\%, the performance of the models generated by the previous version of DENSER (Fast-DENSER) is 86.91\% (statistically significant), which increases to 87.76\% when allowed to train for longer.
\end{abstract}

%
%



%
\maketitle

\section{Introduction}

Automated Machine Learning (AutoML) seeks to model with little or no human-intervention the application of Machine Learning (ML) techniques to well defined problems, avoiding the user to manually perform the data pre-processing, the design and extraction of features, and/or the selection and parameterisation of the most suit ML model. The current paper focuses on a branch of AutoML: NeuroEvolution (NE)~\cite{Floreano2008}. NE automatically searches for Artificial Neural Networks (ANNs), enabling the optimisation of their structure (i.e.., number of neurons, layers, and/or connectivity), and/or learning strategy (i.e., learning algorithm and its parameters: e.g., learning rate, momentum); in NE,  Evolutionary Computation (EC) is used to automate the search for ANNs.

\begin{figure}[t!]
    \scriptsize
    \begin{align}
        {<}\text{fully-connected}{>} ::= & \, \text{layer:fc} \, {<}\text{activation}{>} \\
                    & \, \text{[num-units,int,1,128,2048} \, {<}\text{bias}{>} \\
        {<}\text{dropout}{>} ::= & \text{layer:dropput} \, \text{[rate,float,1,0,0.7]} \\
        {<}\text{activation}{>} ::= & \, \text{act:linear} \, | \, \text{act:relu} \, | \, \text{act:sigmoid}\\
        {<}\text{bias}{>} ::= & \, \text{bias:True} \, | \, \text{bias:False}\\
        {<}\text{softmax}{>} ::= & \, \text{layer:fc} \, \text{act:softmax} \, \text{num-units:10} \, \text{bias:True}\\
        {<}\text{learning}{>} ::= & \, {<}\text{bp}{>} \, \text{[batch\_size,int,1,50,500]} \\
                   & \, | \, {<}\text{rmsprop}{>} \, \text{[batch\_size,int,1,50,500]}  \\
                   & \, | \, {<}\text{adam}{>} \, \text{[batch\_size,int,1,50,500]} \\
         {<}\text{bp}{>} ::= & \, \text{learning:gradient-descent} \, \text{[lr,float,1,0.0001,0.1]}  \\
                   & \, \text{[momentum,float,1,0.68,0.99]} \\
                   & \, \text{[decay,float,1,0.000001,0.001]} \, {<}\text{nesterov}{>} \\
         {<}\text{nesterov}{>} ::= & \, \text{nesterov:True} \, | \, \text{nesterov:False} \\
         {<}\text{adam}{>} ::= & \, \text{learning:adam} \, \text{[lr,float,1,0.0001,0.1]} \\
                               & \, \text{[beta1,float,1,0.5,1]} \, \text{[beta2,float,1,0.5,1]} \\
                               & \, \text{[decay,float,1,0.000001,0.001]} \\
         {<}\text{rmsprop}{>} ::= & \, \text{learning:rmsprop} \, \text{[lr,float,1,0.0001,0.1]} \\
                   & \, \text{[rho,float,1,0.5,1]} \, \text{[decay,float,1,0.000001,0.001]} 
    \end{align}

    \small
    \caption{Example of a grammar for encoding fully-connected networks.}
    \label{fig:grammar_example}
\end{figure}

Considering that NE is based on EC, a population of individuals is continuously evolved throughout generations; each single individual encodes an ANN. One of the main drawbacks relies on the time that is required to evaluate the population, which is even higher if we consider deep networks. To overcome this issue the vast majority of NE methods constraint the evaluation of the networks to a fixed (low) number of epochs, or grant the networks a limited amount of Graphic Processing Unit (GPU) training time. However, these evaluation strategies cannot assure that the candidate solutions are being trained for the required time, and make it difficult to assess the quality of the evolved learning strategy, i.e., that the generated learning strategy works beyond the limited number of epochs / GPU training time.

\begin{figure*}[t!]
    \centering
    \includegraphics[width=.7\textwidth]{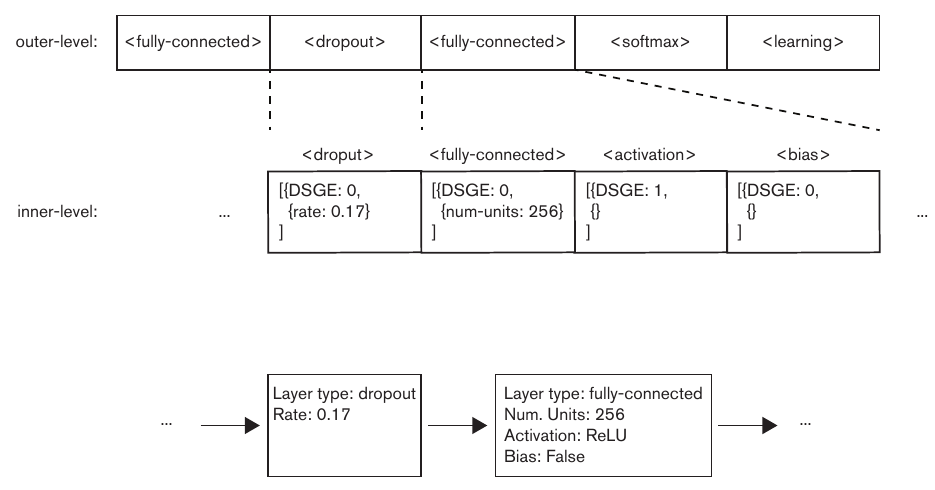}
    \caption{DENSER genotype (top), and respective phenotype (bottom). The example is based on the outer-level structure [((fully-connected, dropout), 1, 10), (softmax, 1, 1), (learning, 1, 1)], and on the grammar of Figure~\ref{fig:grammar_example}.}
    \label{fig:genotype_phenotype_example}
\end{figure*}

To overcome the previous limitation, in this paper we propose a new version of Deep Evolutionary Network Structure Representation (DENSER), called Fast-DENSER++ that enables the evaluation time to grow continuously as the complexity of the networks increases throughout the generations. Fast-DENSER++ (F-DENSER++) is an extension to Fast-DENSER (F-DENSER): a previous version of DENSER that generates networks with the same performance of DENSER, but 20x faster. The results demonstrate that F-DENSER++ is statistically superior to and F-DENSER in evolutionary performance; further, when the F-DENSER networks are trained for longer they achieve, on average, lower performances than those reported by F-DENSER++. That is, F-DENSER++ is able to effectively generate models that are ready for deployment after the end of evolution, without the need for additional training.

The remainder of the paper is organised as follows. Section~\ref{sec:ne} briefly surveys NeuroEvolution works~\ref{sec:ne}. Section~\ref{sec:denser} presents DENSER, and its extension F-DENSER. Section~\ref{sec:f_denser} details F-DENSER++, and the experimental setup and results. Section~\ref{sec:conclusions} draws conclusions and addresses future work.

\section{NeuroEvolution}
\label{sec:ne}

NeuroEvolution (NE) approaches usually focus on the evolution of the learning strategy~\cite{whitley1989applying,Parra2014,DBLP:journals/jmlr/GomezSM08,DBLP:journals/alife/StanleyDG09} or the ~topology~\cite{Gruau:1996:CCE:1595536.1595547,DBLP:conf/icga/MillerTH89}. On the optimisation of the learning strategy NE has been able to match and even surpass the results attained by standard learning algorithms~\cite{DBLP:conf/gecco/MorseS16}; on the other hand, the automatic optimisation of the topology by NE is faster and finds better solutions than using grid or random search~\cite{DBLP:conf/gecco/LorenzoNKRP17}. Nonetheless, when designing a network from the scratch it is hard to separate the learning from the topology, as both are correlated in what regards the search for the most effective model to solve a specific task. Examples of NE approaches that have successfully addressed the simultaneous optimisation of the learning and topology are ~\cite{moriarty2001learning,stanley2002neat,DBLP:conf/gecco/TurnerM13}.

Although the methods are commonly grouped as above, according to the target of evolution, more recent efforts have been put towards the development of methods that are capable of dealing with Deep Artificial Neural Networks (DANNs), and thus we feel that it is more intuitive to divide them into small-scale~\cite{whitley1989applying,stanley2002neat} and large-scale~\cite{DBLP:journals/alife/StanleyDG09,DBLP:journals/corr/MiikkulainenLMR17,DBLP:conf/gecco/SuganumaSN17,DBLP:conf/icml/RealMSSSTLK17,assuncao2018denser} NE. The current paper focuses on the latter; more specifically we will extend F-DENSER~\cite{assunccao2019fast} to enable the generation of models that can be used right-off evolution, without further training. F-DENSER is a general-purpose grammar-based NE approach that can be easily adapted to deal with different problems and/or network types; there is just the need to change the grammar that is feed to the system.

The problem of most of the methods that target the evolution of DANNs is that, even aided by Graphic Processing Units (GPUs) they tend to take a lot of time to find effective models. For example, CoDeepNEAT~\cite{DBLP:journals/corr/MiikkulainenLMR17} train on 100 GPUs, and Real et al. use 450 GPUs for 7 days to perform each run~\cite{real2018regularized}. F-DENSER takes approximately 55 hours (2.3 days) with a single GPU to perform each run, and that is the reason why we have selected F-DENSER for the current paper. There are methods that are computationally cheaper, e.g., Lorenzo and Nalepa~\cite{DBLP:conf/gecco/LorenzoN18} take about 120 minutes to obtain results; however, the speedup is obtained at the cost of the model performance.

\section{Deep Evolutionary Network Structured Representation}
\label{sec:denser}

Deep Evolutionary Network Structured Representation (DENSER)~\cite{assuncao2018denser} is a grammar-based general purpose NE method: it enables the automatic generation of the network topology (sequence, type, connectivity, and parameterisation of the network layers), and learning strategy (learning algorithm and parameterisation). To make this possible DENSER has a two-level representation: (i) the outer-level encodes an ordered sequence of evolutionary units\footnote{The evolutionary units can encode layers, learning strategies, or even data pre-processing and/or augmentation strategies.}; and (ii) the inner-level encodes the parameters of each evolutionary unit. In simple words, each evolutionary unit points to a grammar start symbol, and the grammar itself has every single parameter, and the allowed values. The grammatical nature of DENSER makes the adaption to different network structures and problems easy and transparent, as the user only needs to change the grammar production rules, which are in a text human-readable format. In addition to the grammar, the user needs to define the outer-level structure, which sets the allowed network structure using the following format: [(production-rules, min\_evo\_units, max\_evo\_units), ...]. An example of an outer-level structure for encoding fully-connect networks is [((fully-connected, dropout), 1, 10), (softmax, 1, 1), (learning, 1, 1)], which defines fully-connected networks with between 2 and 11 layers, and a learning block.

Evolution proceeds by a combination of a  Genetic Algorithm (GA) with Dynamic Structured Grammatical Evolution (DSGE). The typical mutation operators of GAs are applied to the outer-level (add, remove, duplicate), and DSGE mutations are applied to the inner-level, i.e., change the expansion possibility, and parameters values. DSGE is chosen over standard Grammatical Evolution (GE) for its ability to deal with the locality and redundancy issues present in GE. To enhance locality the method introduces a one-to-one mapping between the genotype and the non-terminal symbols: there is a list of integers for each non-terminal symbol, and when decoding the genotype the expansion possibility is read from the corresponding list; because each non-terminal symbol has a list associated to it, there is no need for the modulus mathematical operation to select the expansion possibility, and thus redundancy is reduced.

An example of a grammar for encoding fully-connected networks is shown in Figure~\ref{fig:grammar_example}. The  grammar encodes the parameters needed for each evolutionary unit, and are encoded according to the structure: [variable-name, variable-type, num\_values, min\_value, max\_value]. The parameter type can be integer, or float; closed choice parameters are enabled using the grammatical expansion possibilities (e.g., line 5 of the grammar). Figure~\ref{fig:genotype_phenotype_example} represents an example of the genotype and phenotype of an individual using the above outer-level structure, and the grammar of Figure~\ref{fig:grammar_example}.

\begin{algorithm}[t!]
\SetAlgoLined
 parent $\gets$ select\_fittest(population)\\
 
 \eIf{parent.train\_time > DEFAULT\_TIME}{
    tmp\_parent $\gets$ select\_fittest(population-parent)\\
    retrain(tmp\_parent, parent.train\_time)\\
    
    \eIf{tmp\_parent.fitness > parent.fitness}{
        return tmp\_parent
    }{
        return parent
    }
 }{
    return parent
 }
 \caption{Parent selection algorithm.}
 \label{alg:parent_selection}
\end{algorithm}

To speedup search, Fast-DENSER (F-DENSER)~\cite{assunccao2019fast} was introduced: a representation with the same outer and inner-levels is used, and another level is created to encode the connectivity of each layer; this level is referred to as the connectivity-level. Therefore, F-DENSER can evolve not only feed-forward networks but also topologies where any given layer can receive multiple previous layers as input. The same mutation operators are applied to promote evolution, but additionally there are two new operators related to the connectivity-level, that add/remove inputs to layers. In F-DENSER the evolutionary engine is replaced by a (1+$\lambda$)-Evolutionary Strategy (ES). Therefore whilst in DENSER a typically large population of individuals needs to be evaluated, in F-DENSER there is just the need to evaluate (1+$\lambda$) individuals. The results have proved that, without sacrificing performance, F-DENSER with $\lambda=4$ is 20x faster than the original DENSER implementation with a population size of 100 individuals. The previous results are achieved with the same evaluation method, i.e., each individual is trained for a fixed number of 10 epochs. In addition, with the rationale to grant all individuals the same computational resources evolution is conducted with the individuals being trained for a maximum GPU time of 10 minutes; this evaluation stop criteria leads to an improvement of the results.

\begin{table}[t!]
    \centering
    \caption{Experimental parameters.}
    \label{tab:exp_parameters}
    \begin{tabular}{c | c }
        \textbf{Evolutionary Parameter} & \textbf{Value}\\ \hline
        Number of runs & 10 \\ 
        Number of generations & 150\\  
        Population size & 5 \\
        Add layer rate & 25\% \\
        Remove layer rate & 25\% \\
        DSGE-level rate & 15\% \\ 
        & \\
        \textbf{Dataset Parameter} & \textbf{Value} \\ \hline
        Train set & 42500 instances \\
        Validation set & 7500 instances \\
        Test set & 10000 instances \\
        &\\
        \textbf{Train Parameter} & \textbf{Value} \\ \hline
        Default train time & 10 min. \\
        Loss & Categorical Cross-entropy \\
        &\\
        \textbf{Data Augmentation Parameter} & \textbf{Value} \\ \hline
        Padding & 4 \\
        Random crop & 4 \\
        Horizontal flipping & 50\% \\
    \end{tabular}
\end{table}

\section{Fast-DENSER++}
\label{sec:f_denser}

Fast-DENSER++ is an extension to F-DENSER that enables the method to generate networks that are ready for deployment, i.e., the evolutionary result requires no further training to be used. To achieve this we introduce a new mutation operator that does not change any of the layer structure and/or learning parameters, and increases the train time of the individual. Whilst in F-DENSER the maximum train time is set the same for all individuals, in F-DENSER++ the maximum train time is set independently for each individual: in the initial population all individuals are trained for the same amount of time, and the mutation operator changes the maximum train time; any of the other mutation operators reset the evaluation time to the default value, so that the offspring solutions are not evaluated for longer than necessary.

The proposed mutation operator enables the train time to grow continuously as needed, i.e., during the initial generations the networks are simple and thus their train time is reduced, and as time passes more complex solutions require longer evaluations. On the other hand, the new operator makes it possible for individuals within the same population to have different evaluation times. This indirectly implies that the parent selection mechanism has to be changed, so that the comparison between the individuals in the population is fair. In case the fittest individual has been trained for the default train time, the selection is the same as before, i.e., the fittest individual seeds the next generation; otherwise, if the fittest individual is trained for longer than the default train time, the fittest individual of those that were trained for the default train time is re-trained, and the fittest among the two seeds the next generation. That is, the variations of the parent are initially evaluated for the default time, and if in the population there is an individual evaluated for longer, the fittest individual is also granted the same time. The parent selection mechanism is clarified in Algorithm~\ref{alg:parent_selection}. Indirectly we are evolving solutions that have to train fast, but that given more time must improve performance.

To assess the ability of F-DENSER++ to generate ready to deploy DANNs we compare it to the F-DENSER implementation. Therefore we address the generation of Convolutional Neural Networks (CNNs) for the CIFAR-10 dataset. Section~\ref{sec:f_denser_setup} details the experimental setup, and Section~\ref{sec:f_denser_results} the experimental results.

\subsection{Experimental Setup}
\label{sec:f_denser_setup}

\begin{figure}[t!]
    \scriptsize
    \begin{align}
        {<}\text{features}{>} ::= & \, {<}\text{convolution}{>}  \, | \,  {<}\text{convolution}{>}\\
                   & \, | \, {<}\text{pooling}{>} \, | \,  {<}\text{pooling}{>} \\
                   & \, | \, {<}\text{dropout}{>} \, | \, {<}\text{batch-norm}{>} \\
        {<}\text{convolution}{>} ::= & \, \text{layer:conv} \, \text{[num-filters,int,1,32,256]} \, \text{[filter-shape,int,1,2,5]} \\
                    & \, \text{[stride,int,1,1,3]} \, {<}\text{padding}{>} \, {<}\text{activation}{>} \, {<}\text{bias}{>}\\
        {<}\text{batch-norm}{>} ::= & \text{layer:batch-norm}\\
       {<}\text{pooling}{>} ::= & \, {<}\text{pool-type}{>} \, \text{[kernel-size,int,1,2,5]} \\
                    & \, \text{[stride,int,1,1,3]} \, {<}\text{padding}{>} \\
        {<}\text{pool-type}{>} ::= & \, \text{layer:pool-avg} \, | \, \text{layer:pool-max}\\
        {<}\text{padding}{>} ::= & \, \text{padding:same} \, | \, \text{padding:valid}\\
        {<}\text{classification}{>} ::= & \, {<}\text{fully-connected}{>} \, | \,  {<}\text{dropout}{>} \\
        {<}\text{fully-connected}{>} ::= & \, \text{layer:fc} \, {<}\text{activation}{>} \\
                    & \, \text{[num-units,int,1,128,2048} \, {<}\text{bias}{>} \\
        {<}\text{dropout}{>} ::= & \text{layer:dropput} \, \text{[rate,float,1,0,0.7]} \\
        {<}\text{activation}{>} ::= & \, \text{act:linear} \, | \, \text{act:relu} \, | \, \text{act:sigmoid}\\
        {<}\text{bias}{>} ::= & \, \text{bias:True} \, | \, \text{bias:False}\\
        {<}\text{softmax}{>} ::= & \, \text{layer:fc} \, \text{act:softmax} \, \text{num-units:10} \, \text{bias:True}\\
        {<}\text{learning}{>} ::= & \, {<}\text{bp}{>} \, {<}\text{early-stop}{>} \, \text{[batch\_size,int,1,50,500]} \\
                   & \, | \, {<}\text{rmsprop}{>} \, {<}\text{early-stop}{>} \, \text{[batch\_size,int,1,50,500]}  \\
                   & \, | \, {<}\text{adam}{>} \, {<}\text{early-stop}{>} \, \text{[batch\_size,int,1,50,500]} \\
         {<}\text{bp}{>} ::= & \, \text{learning:gradient-descent} \, \text{[lr,float,1,0.0001,0.1]}  \\
                   & \, \text{[momentum,float,1,0.68,0.99]} \, \text{[decay,float,1,0.000001,0.001]} \\
                   & \, {<}\text{nesterov}{>} \\
         {<}\text{nesterov}{>} ::= & \, \text{nesterov:True} \, | \, \text{nesterov:False} \\
         {<}\text{adam}{>} ::= & \, \text{learning:adam} \, \text{[lr,float,1,0.0001,0.1]} \, \text{[beta1,float,1,0.5,1]} \\
                   & \, \text{[beta2,float,1,0.5,1]} \, \text{[decay,float,1,0.000001,0.001]} \\
         {<}\text{rmsprop}{>} ::= & \, \text{learning:rmsprop} \, \text{[lr,float,1,0.0001,0.1]} \\
                   & \, \text{[rho,float,1,0.5,1]} \, \text{[decay,float,1,0.000001,0.001]} \\
         {<}\text{early-stop}{>} ::= & \, \text{[early\_stop,int,1,5,20]}
    \end{align}

    \small
    \caption{Grammar used by F-DENSER++, and F-DENSER for the evolution of CNNs for the CIFAR-10.}
    \label{fig:f_denser_grammar}
\end{figure}

The experimental parameters are detailed in Table~\ref{tab:exp_parameters}, and are divided into 4 sections: (i) evolutionary --  (1+$\lambda$)-ES parameters; (ii) dataset -- number of instances of each of the partitions of the dataset; the dataset is divided into three independent sets: (ii.i) the train set is used for training the individual during evolution; (ii.ii) the validation set is used to perform early stopping, and to assess the fitness of the candidate solution, and (ii.iii) the test is kept out of evolution and used only for assessing the performance of the individuals on unseen data; (iii) train -- fixed training parameters; and (iv) parameters needed for data augmentation.

The reported parameters are the same for F-DENSER++ and F-DENSER, but additionally F-DENSER++ has another mutation rate, that defines the likelihood of an individual to be trained for longer, which is set to 20\%. 

The experiments contained in this section are conducted over the CIFAR-10~\cite{krizhevsky2009learning}: a dataset of 32 $\times$ 32 real-world images of objects. The CIFAR-10 is composed by 50000 train, and 10000 test instances. We search for CNNs for the CIFAR-10 using the grammar of Figure~\ref{fig:f_denser_grammar}, that defines the search space for the topology, and learning strategy.

\subsection{Experimental Results}
\label{sec:f_denser_results}

The results of the evolution of CNNs for the CIFAR-10 with F-DENSER++, and F-DENSER are reported in Table~\ref{tab:f_denser_results}. The evolutionary performance, i.e., fitness (validation), and on unseen data during evolution (test) are summarised; the values are the average of the 10 highest performing networks (according to fitness), one from each run. The analysis of the results makes it clear that F-DENSER++ generates higher performing CNNs than F-DENSER; the variation from the validation to the test set is small, and thus the networks generalise well.

\begin{figure}
    \centering
    \includegraphics[width=.43\textwidth]{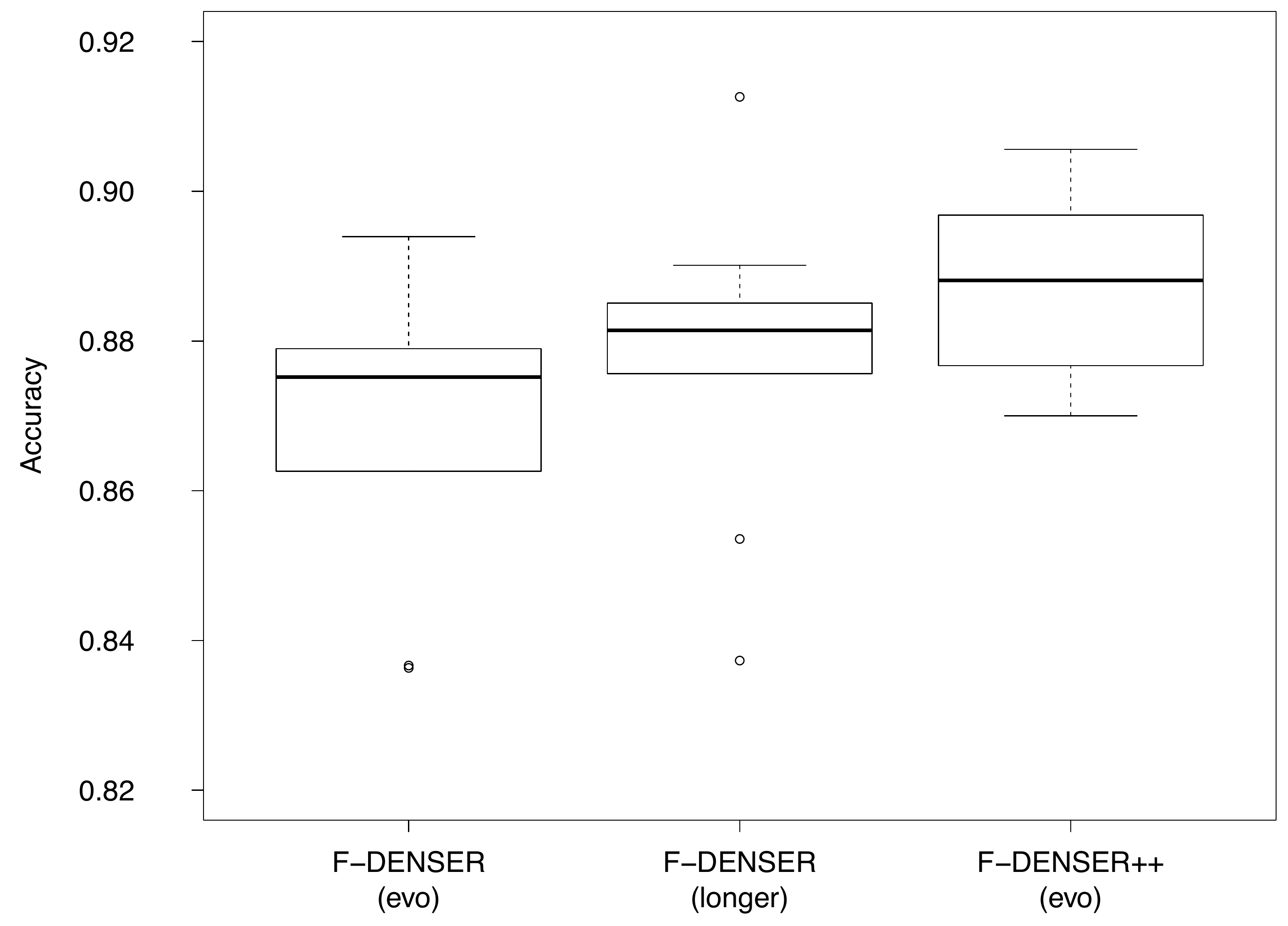}
    \caption{Box-plot of the test accuracies of F-DENSER (evo and longer), and F-DENSER++ (evo).}
    \label{fig:box_plot}
\end{figure}

The significance of the results is tested resorting to statistics. To understand if the samples follow a normal distribution we apply the Kolmogorov-Smirnov and Shapiro-Wilk tests ($\alpha=0.05$). For all the collected data, these tests show that the data does not follow any distribution, and consequently to perform the pairwise comparison we use the Mann-Whitney U test ($\alpha$ = 0.05). In addition, we measure the effect size: low (0.1 $\leq$ r $<$ 0.3), medium (0.3 $\leq$ r $<$ 0.5) and large (r $\geq$ 0.5). The statistical tests show that the difference between F-DENSER++ and F-DENSER is statistically significant, with large effect sizes; the p-values are reported in Table~\ref{tab:f_denser_results}.

In F-DENSER during evolution the networks are evaluated up to a maximum fixed time (10 minutes in the conducted experiments), and thus after evolution the best networks may benefit from re-training for longer. This is not required in F-DENSER++, because during evolution the time allocated for the training of each network can increase. The last row of Table~\ref{tab:f_denser_results} shows the results of F-DENSER, when the networks are re-trained until convergence (determined by early stopping). Even when re-trained the results of F-DENSER++ are slightly superior to those of F-DENSER; however, the difference is not statistically significant.

To better analyse the results we use a box-plot (see Figure~\ref{fig:box_plot}). The plot shows, as above stated, that F-DENSER++ evolutionary results are with no doubts superior to those of F-DENSER. On the other hand, it provides new insights on the comparison (over the test set) of F-DENSER++ with the re-trained networks of F-DENSER: despite not statistically different, the results of F-DENSER++ tend to be superior to the ones reported by F-DENSER -- there are no outliers (despite the slightly larger dispersion), and the median of F-DENSER++ is above the median of F-DENSER; the difference in the median is of approximately 1\%, which translates into about 100 more correctly labeled test instances.

\begin{table}[t!]
    \centering
    \caption{Comparison of the results obtained on the evolution of the topology and learning strategy with F-DENSER++, and F-DENSER on the CIFAR-10. The results report the validation accuracy (fitness), and the test accuracy, and are measured with the generated networks right off evolution (evo), and when trained for longer (longer); the longer training is not applicable to F-DENSER++. The test (longer) results of F-DENSER are compared to the test (evo) results of F-DENSER++. Bold highlights statistically significant results.}
    \label{tab:f_denser_results}
    \begin{tabular}{c|c | c | c }
         & F-DENSER++ & F-DENSER  & p-value \\ \hline
        Validation & 89.44\%  & 87.56\% & \textbf{0.03752} \\ 
        Test (evo) & 88.73\% & 86.91\% & \textbf{0.03156} \\
        Test (longer) & n/a & 87.76\% & 0.30772 \\
    \end{tabular}
\end{table}

From the above, it is demonstrated that F-DENSER++ can effectively generate networks that are ready to be deployed right-off evolution, i.e., there is no need for further training. This helps in the testing of the evolved training policy, as it is used until convergence; the training policies that are generated for F-DENSER despite providing good results when applied for longer training cycles may not be the most adequate ones. Most importantly, the above results are achieved without a major increase in the time required to search for the networks: from an average of 0.73 hours/generation to an average of 1.13 hours/generation, which is still fairly bellow the average of 10.83 hours/generation of DENSER\footnote{All the times are measure in machines with the same specifications:  1080 Ti GPUs, 64 GB of RAM, and an Intel Core i7-6850K CPU.}. From this point onward we focus on the use of F-DENSER++.

\section{Conclusions}
\label{sec:conclusions}

The current work introduces F-DENSER++: an extension to F-DENSER that enables it to generate fully-trained models, i.e., models that can be deployed right-off evolution. The results demonstrate that the evolutionary results of F-DENSER++ are statistically superior to those of F-DENSER. The results of F-DENSER still need to be trained for longer after evolution; nonetheless, the performance of the longer trains is still bellow the evolutionary performance of F-DENSER++. In addition, we can state that the new method is superior to the standard DENSER implementation; the evolutionary results of F-DENSER are statistically superior to DENSER, and F-DENSER++ is statistically superior to F-DENSER, and consequently superior to DENSER.

Future work  will be guided into two separate directions: (i) perform experiments with a wider set of datasets, and (ii) investigate transfer and multi-task learning with F-DENSER++. The common approach to NE seeks to generate a network for a specific task, without using any of the information gathered when addressing previous tasks. In the future, it is our objective to evolve DENSER to a point where learning is incremental and cumulative, using past knowledge, and avoiding catastrophic forgetting. That is, we want a system that grows with time, and learns new tasks without stopping being able to solve the previous ones.

\begin{acks}

The work is partially supported by the \grantsponsor{SP4970}{Portuguese Foundation for Science and Technology}{http://fct.pt} under Grant No.:~\grantnum{SP4970}{SFRH/BD/114865/2016}.

\end{acks}

\bibliographystyle{ACM-Reference-Format}
\bibliography{sample-base}


\begin{thebibliography}{20}


\ifx \showCODEN    \undefined \def \showCODEN     #1{\unskip}     \fi
\ifx \showDOI      \undefined \def \showDOI       #1{#1}\fi
\ifx \showISBNx    \undefined \def \showISBNx     #1{\unskip}     \fi
\ifx \showISBNxiii \undefined \def \showISBNxiii  #1{\unskip}     \fi
\ifx \showISSN     \undefined \def \showISSN      #1{\unskip}     \fi
\ifx \showLCCN     \undefined \def \showLCCN      #1{\unskip}     \fi
\ifx \shownote     \undefined \def \shownote      #1{#1}          \fi
\ifx \showarticletitle \undefined \def \showarticletitle #1{#1}   \fi
\ifx \showURL      \undefined \def \showURL       {\relax}        \fi
\providecommand\bibfield[2]{#2}
\providecommand\bibinfo[2]{#2}
\providecommand\natexlab[1]{#1}
\providecommand\showeprint[2][]{arXiv:#2}

\bibitem[\protect\citeauthoryear{Assun{\c{c}}{\~a}o, Louren{\c{c}}o, Machado,
  and Ribeiro}{Assun{\c{c}}{\~a}o et~al\mbox{.}}{2018}]%
        {assuncao2018denser}
\bibfield{author}{\bibinfo{person}{Filipe Assun{\c{c}}{\~a}o},
  \bibinfo{person}{Nuno Louren{\c{c}}o}, \bibinfo{person}{Penousal Machado},
  {and} \bibinfo{person}{Bernardete Ribeiro}.} \bibinfo{year}{2018}\natexlab{}.
\newblock \showarticletitle{DENSER: deep evolutionary network structured
  representation}.
\newblock \bibinfo{journal}{\emph{Genetic Programming and Evolvable Machines}}
  (\bibinfo{date}{27 Sep} \bibinfo{year}{2018}).
\newblock
\showISSN{1573-7632}
\urldef\tempurl%
\url{https://doi.org/10.1007/s10710-018-9339-y}
\showDOI{\tempurl}


\bibitem[\protect\citeauthoryear{Assun{\c{c}}{\~a}o, Louren{\c{c}}o, Machado,
  and Ribeiro}{Assun{\c{c}}{\~a}o et~al\mbox{.}}{2019}]%
        {assunccao2019fast}
\bibfield{author}{\bibinfo{person}{Filipe Assun{\c{c}}{\~a}o},
  \bibinfo{person}{Nuno Louren{\c{c}}o}, \bibinfo{person}{Penousal Machado},
  {and} \bibinfo{person}{Bernardete Ribeiro}.} \bibinfo{year}{2019}\natexlab{}.
\newblock \showarticletitle{Fast DENSER: Efficient Deep NeuroEvolution}. In
  \bibinfo{booktitle}{\emph{European Conference on Genetic Programming}}.
  Springer, \bibinfo{pages}{197--212}.
\newblock


\bibitem[\protect\citeauthoryear{Floreano, D{\"u}rr, and Mattiussi}{Floreano
  et~al\mbox{.}}{2008}]%
        {Floreano2008}
\bibfield{author}{\bibinfo{person}{Dario Floreano}, \bibinfo{person}{Peter
  D{\"u}rr}, {and} \bibinfo{person}{Claudio Mattiussi}.}
  \bibinfo{year}{2008}\natexlab{}.
\newblock \showarticletitle{Neuroevolution: from architectures to learning}.
\newblock \bibinfo{journal}{\emph{Evolutionary Intelligence}}
  \bibinfo{volume}{1}, \bibinfo{number}{1} (\bibinfo{date}{01 Mar}
  \bibinfo{year}{2008}), \bibinfo{pages}{47--62}.
\newblock
\showISSN{1864-5917}
\urldef\tempurl%
\url{https://doi.org/10.1007/s12065-007-0002-4}
\showDOI{\tempurl}


\bibitem[\protect\citeauthoryear{Gomez, Schmidhuber, and Miikkulainen}{Gomez
  et~al\mbox{.}}{2008}]%
        {DBLP:journals/jmlr/GomezSM08}
\bibfield{author}{\bibinfo{person}{Faustino~J. Gomez},
  \bibinfo{person}{J{\"{u}}rgen Schmidhuber}, {and} \bibinfo{person}{Risto
  Miikkulainen}.} \bibinfo{year}{2008}\natexlab{}.
\newblock \showarticletitle{Accelerated Neural Evolution through Cooperatively
  Coevolved Synapses}.
\newblock \bibinfo{journal}{\emph{Journal of Machine Learning Research}}
  \bibinfo{volume}{9} (\bibinfo{year}{2008}), \bibinfo{pages}{937--965}.
\newblock


\bibitem[\protect\citeauthoryear{Gruau, Whitley, and Pyeatt}{Gruau
  et~al\mbox{.}}{1996}]%
        {Gruau:1996:CCE:1595536.1595547}
\bibfield{author}{\bibinfo{person}{Fr{\'e}d{\'e}ric Gruau},
  \bibinfo{person}{Darrell Whitley}, {and} \bibinfo{person}{Larry Pyeatt}.}
  \bibinfo{year}{1996}\natexlab{}.
\newblock \showarticletitle{A Comparison Between Cellular Encoding and Direct
  Encoding for Genetic Neural Networks}. In
  \bibinfo{booktitle}{\emph{Proceedings of the 1st Annual Conference on Genetic
  Programming}}. \bibinfo{publisher}{MIT Press}, \bibinfo{address}{Cambridge,
  MA, USA}, \bibinfo{pages}{81--89}.
\newblock
\showISBNx{0-262-61127-9}
\urldef\tempurl%
\url{http://dl.acm.org/citation.cfm?id=1595536.1595547}
\showURL{%
\tempurl}


\bibitem[\protect\citeauthoryear{Krizhevsky and Hinton}{Krizhevsky and
  Hinton}{2009}]%
        {krizhevsky2009learning}
\bibfield{author}{\bibinfo{person}{Alex Krizhevsky} {and}
  \bibinfo{person}{Geoffrey Hinton}.} \bibinfo{year}{2009}\natexlab{}.
\newblock \bibinfo{booktitle}{\emph{Learning multiple layers of features from
  tiny images}}.
\newblock \bibinfo{type}{{T}echnical {R}eport}.
  \bibinfo{institution}{Citeseer}.
\newblock


\bibitem[\protect\citeauthoryear{Lorenzo and Nalepa}{Lorenzo and
  Nalepa}{2018}]%
        {DBLP:conf/gecco/LorenzoN18}
\bibfield{author}{\bibinfo{person}{Pablo~Ribalta Lorenzo} {and}
  \bibinfo{person}{Jakub Nalepa}.} \bibinfo{year}{2018}\natexlab{}.
\newblock \showarticletitle{Memetic evolution of deep neural networks}. In
  \bibinfo{booktitle}{\emph{{GECCO}}}. \bibinfo{publisher}{{ACM}},
  \bibinfo{pages}{505--512}.
\newblock


\bibitem[\protect\citeauthoryear{Lorenzo, Nalepa, Kawulok, Ramos, and
  Pastor}{Lorenzo et~al\mbox{.}}{2017}]%
        {DBLP:conf/gecco/LorenzoNKRP17}
\bibfield{author}{\bibinfo{person}{Pablo~Ribalta Lorenzo},
  \bibinfo{person}{Jakub Nalepa}, \bibinfo{person}{Michal Kawulok},
  \bibinfo{person}{Luciano~S{\'{a}}nchez Ramos}, {and}
  \bibinfo{person}{Jos{\'{e}}~Ranilla Pastor}.}
  \bibinfo{year}{2017}\natexlab{}.
\newblock \showarticletitle{Particle swarm optimization for hyper-parameter
  selection in deep neural networks}. In \bibinfo{booktitle}{\emph{{GECCO}}}.
  \bibinfo{publisher}{{ACM}}, \bibinfo{pages}{481--488}.
\newblock


\bibitem[\protect\citeauthoryear{Miikkulainen, Liang, Meyerson, Rawal, Fink,
  Francon, Raju, Shahrzad, Navruzyan, Duffy, and Hodjat}{Miikkulainen
  et~al\mbox{.}}{2017}]%
        {DBLP:journals/corr/MiikkulainenLMR17}
\bibfield{author}{\bibinfo{person}{Risto Miikkulainen},
  \bibinfo{person}{Jason~Zhi Liang}, \bibinfo{person}{Elliot Meyerson},
  \bibinfo{person}{Aditya Rawal}, \bibinfo{person}{Daniel Fink},
  \bibinfo{person}{Olivier Francon}, \bibinfo{person}{Bala Raju},
  \bibinfo{person}{Hormoz Shahrzad}, \bibinfo{person}{Arshak Navruzyan},
  \bibinfo{person}{Nigel Duffy}, {and} \bibinfo{person}{Babak Hodjat}.}
  \bibinfo{year}{2017}\natexlab{}.
\newblock \showarticletitle{Evolving Deep Neural Networks}.
\newblock \bibinfo{journal}{\emph{CoRR}}  \bibinfo{volume}{abs/1703.00548}
  (\bibinfo{year}{2017}).
\newblock


\bibitem[\protect\citeauthoryear{Miller, Todd, and Hegde}{Miller
  et~al\mbox{.}}{1989}]%
        {DBLP:conf/icga/MillerTH89}
\bibfield{author}{\bibinfo{person}{Geoffrey~F. Miller},
  \bibinfo{person}{Peter~M. Todd}, {and} \bibinfo{person}{Shailesh~U. Hegde}.}
  \bibinfo{year}{1989}\natexlab{}.
\newblock \showarticletitle{Designing Neural Networks using Genetic
  Algorithms}. In \bibinfo{booktitle}{\emph{{ICGA}}}.
  \bibinfo{publisher}{Morgan Kaufmann}, \bibinfo{pages}{379--384}.
\newblock


\bibitem[\protect\citeauthoryear{Moriarty and Miikkulainen}{Moriarty and
  Miikkulainen}{2001}]%
        {moriarty2001learning}
\bibfield{author}{\bibinfo{person}{David~E Moriarty} {and}
  \bibinfo{person}{Risto Miikkulainen}.} \bibinfo{year}{2001}\natexlab{}.
\newblock \showarticletitle{Learning sequential decision tasks through
  symbiotic evolution of neural networks}.
\newblock \bibinfo{journal}{\emph{Advances in the Evolutionary Synthesis of
  Intelligent Agents}} (\bibinfo{year}{2001}), \bibinfo{pages}{367}.
\newblock


\bibitem[\protect\citeauthoryear{Morse and Stanley}{Morse and Stanley}{2016}]%
        {DBLP:conf/gecco/MorseS16}
\bibfield{author}{\bibinfo{person}{Gregory Morse} {and}
  \bibinfo{person}{Kenneth~O. Stanley}.} \bibinfo{year}{2016}\natexlab{}.
\newblock \showarticletitle{Simple Evolutionary Optimization Can Rival
  Stochastic Gradient Descent in Neural Networks}. In
  \bibinfo{booktitle}{\emph{{GECCO}}}. \bibinfo{publisher}{{ACM}},
  \bibinfo{pages}{477--484}.
\newblock


\bibitem[\protect\citeauthoryear{Parra, Trujillo, and Melin}{Parra
  et~al\mbox{.}}{2014}]%
        {Parra2014}
\bibfield{author}{\bibinfo{person}{Jos{\'e} Parra}, \bibinfo{person}{Leonardo
  Trujillo}, {and} \bibinfo{person}{Patricia Melin}.}
  \bibinfo{year}{2014}\natexlab{}.
\newblock \showarticletitle{Hybrid back-propagation training with evolutionary
  strategies}.
\newblock \bibinfo{journal}{\emph{Soft Computing}} \bibinfo{volume}{18},
  \bibinfo{number}{8} (\bibinfo{date}{01 Aug} \bibinfo{year}{2014}),
  \bibinfo{pages}{1603--1614}.
\newblock
\showISSN{1433-7479}
\urldef\tempurl%
\url{https://doi.org/10.1007/s00500-013-1166-8}
\showDOI{\tempurl}


\bibitem[\protect\citeauthoryear{Real, Aggarwal, Huang, and Le}{Real
  et~al\mbox{.}}{2018}]%
        {real2018regularized}
\bibfield{author}{\bibinfo{person}{Esteban Real}, \bibinfo{person}{Alok
  Aggarwal}, \bibinfo{person}{Yanping Huang}, {and} \bibinfo{person}{Quoc~V
  Le}.} \bibinfo{year}{2018}\natexlab{}.
\newblock \showarticletitle{Regularized evolution for image classifier
  architecture search}.
\newblock \bibinfo{journal}{\emph{arXiv preprint arXiv:1802.01548}}
  (\bibinfo{year}{2018}).
\newblock


\bibitem[\protect\citeauthoryear{Real, Moore, Selle, Saxena, Suematsu, Tan, Le,
  and Kurakin}{Real et~al\mbox{.}}{2017}]%
        {DBLP:conf/icml/RealMSSSTLK17}
\bibfield{author}{\bibinfo{person}{Esteban Real}, \bibinfo{person}{Sherry
  Moore}, \bibinfo{person}{Andrew Selle}, \bibinfo{person}{Saurabh Saxena},
  \bibinfo{person}{Yutaka~Leon Suematsu}, \bibinfo{person}{Jie Tan},
  \bibinfo{person}{Quoc~V. Le}, {and} \bibinfo{person}{Alexey Kurakin}.}
  \bibinfo{year}{2017}\natexlab{}.
\newblock \showarticletitle{Large-Scale Evolution of Image Classifiers}. In
  \bibinfo{booktitle}{\emph{{ICML}}} \emph{(\bibinfo{series}{Proceedings of
  Machine Learning Research})}, Vol.~\bibinfo{volume}{70}.
  \bibinfo{publisher}{{PMLR}}, \bibinfo{pages}{2902--2911}.
\newblock


\bibitem[\protect\citeauthoryear{Stanley, D'Ambrosio, and Gauci}{Stanley
  et~al\mbox{.}}{2009}]%
        {DBLP:journals/alife/StanleyDG09}
\bibfield{author}{\bibinfo{person}{Kenneth~O. Stanley},
  \bibinfo{person}{David~B. D'Ambrosio}, {and} \bibinfo{person}{Jason Gauci}.}
  \bibinfo{year}{2009}\natexlab{}.
\newblock \showarticletitle{A Hypercube-Based Encoding for Evolving Large-Scale
  Neural Networks}.
\newblock \bibinfo{journal}{\emph{Artificial Life}} \bibinfo{volume}{15},
  \bibinfo{number}{2} (\bibinfo{year}{2009}), \bibinfo{pages}{185--212}.
\newblock


\bibitem[\protect\citeauthoryear{Stanley and Miikkulainen}{Stanley and
  Miikkulainen}{2002}]%
        {stanley2002neat}
\bibfield{author}{\bibinfo{person}{Kenneth~O. Stanley} {and}
  \bibinfo{person}{Risto Miikkulainen}.} \bibinfo{year}{2002}\natexlab{}.
\newblock \showarticletitle{Evolving Neural Networks Through Augmenting
  Topologies}.
\newblock \bibinfo{journal}{\emph{Evol. Comput.}} \bibinfo{volume}{10},
  \bibinfo{number}{2} (\bibinfo{date}{June} \bibinfo{year}{2002}),
  \bibinfo{pages}{99--127}.
\newblock
\showISSN{1063-6560}
\urldef\tempurl%
\url{https://doi.org/10.1162/106365602320169811}
\showDOI{\tempurl}


\bibitem[\protect\citeauthoryear{Suganuma, Shirakawa, and Nagao}{Suganuma
  et~al\mbox{.}}{2017}]%
        {DBLP:conf/gecco/SuganumaSN17}
\bibfield{author}{\bibinfo{person}{Masanori Suganuma},
  \bibinfo{person}{Shinichi Shirakawa}, {and} \bibinfo{person}{Tomoharu
  Nagao}.} \bibinfo{year}{2017}\natexlab{}.
\newblock \showarticletitle{A genetic programming approach to designing
  convolutional neural network architectures}. In
  \bibinfo{booktitle}{\emph{{GECCO}}}. \bibinfo{publisher}{{ACM}},
  \bibinfo{pages}{497--504}.
\newblock


\bibitem[\protect\citeauthoryear{Turner and Miller}{Turner and Miller}{2013}]%
        {DBLP:conf/gecco/TurnerM13}
\bibfield{author}{\bibinfo{person}{Andrew~James Turner} {and}
  \bibinfo{person}{Julian~Francis Miller}.} \bibinfo{year}{2013}\natexlab{}.
\newblock \showarticletitle{Cartesian genetic programming encoded artificial
  neural networks: a comparison using three benchmarks}. In
  \bibinfo{booktitle}{\emph{{GECCO}}}. \bibinfo{publisher}{{ACM}},
  \bibinfo{pages}{1005--1012}.
\newblock


\bibitem[\protect\citeauthoryear{Whitley}{Whitley}{1989}]%
        {whitley1989applying}
\bibfield{author}{\bibinfo{person}{Darrell Whitley}.}
  \bibinfo{year}{1989}\natexlab{}.
\newblock \showarticletitle{Applying genetic algorithms to neural network
  learning}. In \bibinfo{booktitle}{\emph{Proceedings of the Seventh Conference
  (AISB89) on Artificial Intelligence and Simulation of Behaviour}}. Morgan
  Kaufmann Publishers Inc., \bibinfo{pages}{137--144}.
\newblock


\end{thebibliography}

\end{document}